**HTM-MAT: An online prediction software toolbox based on cortical machine learning algorithm**

V.I.E Anireh[a], E.N. Osegi [b,*]

   a. Department of Computer Science, Rivers State University of Science and Technology, Nigeria
   b. Department of Information and Communication Technology, National Open University of Nigeria

**Abstract.**
HTM-MAT is a MATLAB® - based toolbox for implementing cortical learning algorithms (CLA) including related cortical-like algorithms that possesses spatiotemporal properties. CLA is a suite of predictive machine learning algorithms developed by Numenta Inc. and is based on the hierarchical temporal memory (HTM). This paper presents an implementation of HTM-MAT with several illustrative examples including several toy datasets and compared with two sequence learning applications employing state-of-the-art algorithms - the recurrentjs based on the Long Short-Term Memory (LSTM) algorithm and OS-ELM which is based on an online sequential version of the Extreme Learning Machine. The performance of HTM-MAT using two historical benchmark datasets and one real world dataset is also compared with one of the existing sequence learning applications, the OS-ELM. The results indicate that HTM-MAT predictions are indeed competitive and can outperform OS-ELM in sequential prediction tasks.

**Keywords:**
*Machine learning; predictions; cortical learning algorithms; hierarchical temporal memory*

*Corresponding author.
E-mail:* nd.osegi@sure-gp.com

## 1. Motivation and Significance

Predictive systems are very vital in the monitoring and operation of modern day businesses, infrastructure and equipment's. For instance in the power industry, utility operators need to determine the expected load profile in advance in order to make proper load schedule for the day; bank operators need to determine the likelihood that a fraudulent bank transaction will occur; web system operators need to detect rogue like behaviour in order to avert security breaches within a given web portal. In order to gain insight into the behaviour of the aforementioned tasks, such systems require predictive coding techniques which in turn play a major role in predicting the hidden causes of incoming sensory information [1]. In recent times, cortical learning algorithms (CLA) have been proposed in [2] and also applied to the solution of several real challenging problems including retina analysis [3], sign language and gaze gesture recognition [4-5] and weld flaw detection [6] have been investigated with very promising results. This suite of algorithms has very interesting properties and represents a major step in the quest for a more biological plausible neural network for machine learning tasks. However, until this present moment, an open source cortical learning toolbox that can facilitate research and understanding of CLA-like algorithms in a variety of machine learning predictive tasks is lacking in the technical computing community.

In this paper we present HTM-MAT, a software tool which allows such predictive tasks to be accomplished. HTM-MAT is a barebones minimalist version of the cortical learning algorithm proposed in [2] that allows regression and classification functions to be implemented in a predictive way. The theory is based on the hypothesis of the use of spatiotemporal sparse memory predictive neural architecture for online machine learning tasks. The primary purpose of developing HTM-MAT is to make readily available a domain friendly, reusable and comprehensive cortical-learning toolbox for machine learning experiments to be performed on both synthetic and real world data in a technical computing environment.

It is intended to serve as a reusable, straightforward and easy to re-invent/reuse machine learning tool for academic researchers, students and machine learning experts. HTM-MAT depends on the Hierarchical Temporal Memory (HTM) which is the underlying theory that defines its core operations.

### 1.1. HTM theoretical background

Hierarchical Temporal Memory (HTM) is originally a memory-prediction framework, and an emerging Artificial Intelligence (AI) approach that is based on the notion that time plays a crucial role in the operation of neural networks [2,7-9]. In its current implementation, it uses the cortical learning algorithms - a suite of algorithms that favors the use of a more biological plausible paradigm. HTM advocates that the neocortex is the primary entity and the main seat of intelligence in the brain.

The core operations that make HTM interestingly unique are centered on the properties of sparse distributed representations (SDRs) being the fundamental data structure in HTM systems [10]. The primary operations in HTM include [9]:

- The activation and prediction of cell states/columns based on the notion of permanence
- Computation of a similarity metric called the "overlap"
- A learning rule for updating permanence's

The prediction and activation states at time step, t, are typically computed using Eq. (1) and Eq. (2) respectively as:

$$\pi_{ij}^t = \begin{cases} 1 & if \ \exists_d \ \|\tilde{D}_{ij}^d \circ A^t\|_1 > \theta \\ 0 & otherwise \end{cases} \quad (1)$$

$$a_{ij}^t = \begin{cases} 1 & if\ j \in W^t\ and\ \pi_{ij}^{t-1} = 1 \\ 1 & if\ j \in W^t\ and\ \sum_i \pi_{ij}^{t-1} = 0 \\ 0 & otherwise \end{cases} \qquad (2)$$

where,

$\tilde{D}_{ij}^d$ = an M×N binary matrix representing the permanence of a connected synapse belonging to a dendrite segment

$d$ = a cortical segment

$i, j$ = cell and column states

$A^t$ = an M×N binary matrix denoting the activation state of the cortical network
$\theta$ = the cortical segment activation threshold or permanence
$N$ = number of cortical columns
$M$ = number of neurons (cortical cells) per column
$W^t$ = percentage of columns with most synaptic inputs

The operation $\|\tilde{D}_{ij}^d \circ A^t\|_1$, is fundamentally referred to as the "overlap".

In order to improve on the HTM predictions, a Hebbian-like learning rule is applied to the dendrite segment:

$$\Delta D_{ij}^d = p^+ \left( D_{ij}^d \circ A^{t-1} \right) - p^- D_{ij}^d \qquad (3)$$

where, $p^+$ and $p^-$ are permanence increments and decrements respectively.

The percentage of active columns typically follows a 2% rule for maximum fixed sparse distribution. The assumption that the 2% rule holds and can be determined apriori remains a fundamental issue and yet to be resolved by HTM researchers. More so, it is still not clear if the brain uses this rule for learning sparsely. Nonetheless, the HTM model still works reasonably well in certain sequence learning tasks such as in time series related problems.

*1.2. Concept of an online sequential cortical learning algorithm*

*Online sequential learning emphasizes the predictions of patterns that show some interesting regularities or properties in data e.g. the prediction of the likelihood of an anomaly in equipment given a history of equipment conditions. HTM-MAT is a sequence learning framework based on cortical learning principles and is generically designed to operate in an online manner.*

*A typical online learning process for cortical-like algorithms consists of four key stages. First, a dataset containing the source file (sensory input data) is encoded into a suitable (mixed-integer) representation. This is achieved by using a scalar encoder. Second, using the concepts introduced earlier in Section 1.1, a sparse distributed representation (SDR) of the encoder representations is generated using a spatial pooler algorithm. Third, the SDR formed in the spatial pooler stage is further processed to generate a hierarchy of sparse representations using a temporal pooler algorithm. Fourth, a greedy prediction search is performed to extract the possible sparse pattern meeting a pre-specified overlap threshold. Full details of the spatiotemporal processing inherent in cortical-like applications including a comprehensive mathematical treatment of the formulation of sparse distributed structures can be found in [2, 9 and 10]. Note, that in HTM-MAT, we do not consider the expensive method of combinatorics proposed in [10], but rather use a Monte Carlo simulation to evolve a sparse distributed representation. An implementation of HTM-MAT is provided in [11].*

*1.2.1 Related sequential learning algorithms*

*The online sequential extreme learning machine (OS-ELM) is also an emerging sequence learning framework that is based on the very popular Extreme Learning Machines (ELM).OS-ELM has been shown to be very fast and highly accurate [12]. However, this model does not fully account for the notion of time and hierarchy in space. It also does not account for sparse distributed representations which are core features of truly biological-plausible machine learning applications. In addition OS-ELM does not have any inherent mechanism for handling categorical inputs and requires supervised operation. Implementations of OS-ELM can be found in [13].*

*Another sequence learning algorithm, Long Short-Term Memory (LSTM) neural network originally developed in [14] represent the state-of-the-art in predictive mining tasks specifically in the area of language modelling and grammar recognition/prediction. It can operate both in supervised, semi-supervised and unsupervised modes. In particular, LSTM networks can be used as a reference benchmark for evaluating the predictive power of other machine learning algorithms or neural networks where the task is to predict in advance, the possible output response state given a history of previous states. LSTM model neural networks have been used in a variety of tasks including language modelling [15-17], grammar/word learning tasks [18] and speech recognition [19, 20]. An implementation of LSTM can be found in [21].*

*1.3. HTM-MAT*

*HTM-MAT is a MATLAB ® toolbox for performing machine learning experiments on streaming or sequential data. The primary feature of HTM-MAT is that it uses a sparse mixed-integer representation for encoding data during training and learning. Using a sparse mixed integer representation gives HTM-MAT the advantage of a character-level or symbolic explanation facility i.e. HTM-MAT can generate more meaningful and expressive results that still follow a sparse distributed representation using the core HTM principles.*

*Since HTM-MAT uses an unsupervised machine learning algorithm, it can also derive a solution to the problem in a more compact and adaptive manner. Though, HTM-MAT is specifically intended for researchers and students in academic environments, it may be possible to adapt it to industry or commercial applications.HTM-MAT can also be helpful to users new to the use of cortical-like machine learning algorithms for predictive classification tasks.*

## 2. Software description

HTM-MAT exploits the functional programming capabilities of MATLAB ®. HTM-MAT interacts with the user through a set of functional routines typically instantiated in a main functional class. The architectural concept is as shown in Fig.1. Basically, the user interacts with HTM-MAT (top-layer) through a set of functions (middle-layer). These functions include routines for handling spatial and temporal pooling operations as well as for classifying the recognition units. The lower layer (DATASET FILE) is used to define the dataset needed for the prediction or classification task.

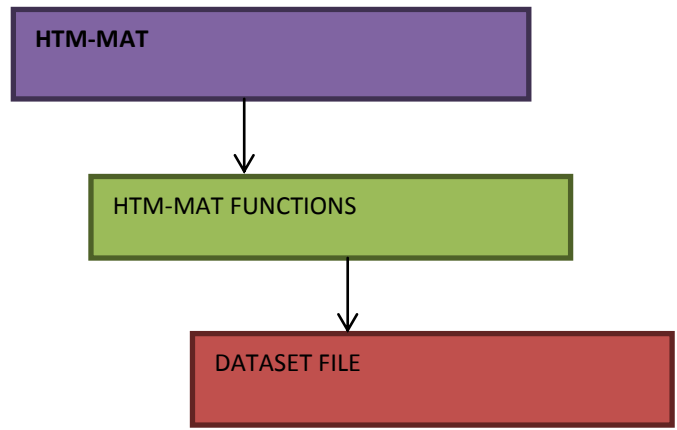

Fig1.Architecture of HTM-MAT application

Full details of the core functions used in HTM-MAT can be found in the user guide that comes with the software package [11].

## 3. Illustrative Examples

A sample implementation of a prediction task using a synthetic or benchmark dataset is as shown in Table 1. In this task, the goal is to implement a code routine to predict the target value of the last exemplar given a history of previous inputs. This routine consists of the following blocks of code:

- HTM-System Parameters: First the data source (dataset) name is defined followed by a definition of several core system parameters such as the number of Monte Carlo (MC) iterations (iters), minimum overlap (min_overlap), permanence threshold (perms_th), the number of considered sequences (seq_size) and the percent overlap adjustment for greedy-max predictions.
- Spatiotemporal processing: This requires a function call to ST_HTM.m functional class using most of the parameters defined in the HTM-System Parameters block. This operation returns the full mean absolute percentage error (MAPE), a mixed-integer representation of the original input data sequence (original_data) and sparse generated data (sparsed_data), the overlap (overlap), temporal sparse data (sparsed_data_T), minimum values obtained meeting the overlap threshold (min_val) and the permanence.
- Error-plots: Plots the MAPE errors generated by the system after the pre-specified number of MC iterations.
- Char-Integer transformation: This transformation converts the mixed-integer character representations of original and sparse generated data into an equivalent numeric (integer-based) representation.This is very important for extracting the predictions (using a prefix-search) and performing the necessary

- *classifications.*
- *Argmax-predictions: This block allows greedy character-wise predictions to be performed on the recognition units. The predictions are stored in a variable called predicted_out.*
- *Performance metrics: This block computes the performance metrics such as the percentage classification accuracy, mean absolute percentage error (MAPE), and the root-mean squared error (RMSE) given the original (in_p) and sparse recognition (out_s) integer units.*

*Table1. A sample HTM-MAT implementation*

```
data_name = 'heart_data'; % dataset
%% HTM system parameters:
iters = 50;
min_overlap = 2;
perms_th = 0.21;
desired_localActivity = 2;
seq_size = 700;
per_adjust = 99;
%% Spatiotemporal processing:
[MAPE_n,original_data,sparsed_data,overlap,sparsed_data_T,min_val,permanence] = ST_HTM(iters,...
    min_overlap,perms_th,desired_localActivity,data_name,seq_size);
%% Error Plots:
plot(MAPE_n);
%% Char-Integer Transformation:
in_p = original_data.*1;
out_s = sparsed_data.*1;
%% Argmax-predictions:
        predicted_out(pred_idx_n,:) = char(out_s(pred_idx_n,:));
%% Performance metrics:
%Compute the errors:
RMSE = RMSE_func(in_p,out_s)
```

*In Table 2 is shown a comparative prediction of the HTM-MAT with two applications employing state-of-the-art algorithms using four toy datasets; these applications, recurrentjs developed in [21] and OS-ELM applications using three different activation functions [13] have been earlier introduced in sub-section 1.2.1. The toy datasets include a multiplication operation obtained from the very popular times-table used in high-school mathematics, and three fictitious datasets of word representations (applicable to only HTM-MAT and recurrentjs since they both can handle character-level input data). These datasets are included in the HTM-MAT toolbox [11].*

*In the first toy dataset (times_trainv1), the target value of the last exemplar was intentionally set to zero in order to find out if the algorithm can accurately compute the correct answer.*

*Table 3 also shows comparative experimental results of HTM-MAT with the OS-ELM applications using two historical benchmark datasets and one field data for the predictive classification task. The historical datasets can be obtained from [23] and consists of a heart dataset (heart_data) for categorizing a heart condition as either present (1) or absent (0) and a fraud dataset (australian_data) for detecting the presence (1) or absence of a fraud (0) . The field data represent two-hourly pressure readings obtained from a sensor attached to an oil pipeline where a 1 indicates a likely threat on the facility. These results are interpreted in terms of the root-mean-squared error*

(RMSE). Simulations were performed for a maximum of 50 MC iterations for HTM-MAT. The results in Table 2 indicate that HTM-MAT is indeed capable of numeric and word predictions that are indeed a reflection of the pattern in the dataset. The results in Table 3 show that HTM-MAT outperforms the OS-ELM on all tasks presented and achieves 100% accuracy on the real world dataset.

It is important to note that the OS-ELM parameters was especially hand-tuned to obtain the best possible set of predictions which was found to be achievable when the number of hidden neurons is greater than 100 and the number of initial data and block size is set to 3 and 0 respectively.

Table2. Comparative word/numeric predictions of HTM-MAT, recurrentjs and OS-ELM using toy datasets

| Dataset | recurrentjs | OS-ELMrbf [a] | OS-ELMsig [b] | OS-ELMsin [c] | HTM-MAT |
|---|---|---|---|---|---|
| times_trainv1 | 2 3 6 | 2 3 6.1 | 2 3 5.5 | 2 3 5.8 | 2 3 6 |
| word3a | Fishing, Fish-feed, Fish | NA | NA | NA | Fast, F sh |
| word3b | Football, Fans | NA | NA | NA | Foot |
| word3c | Video-Player, Video | NA | NA | NA | Video |

[a] OS-ELMrbf – OS-ELM using radial basis activation function
[b] OS-ELMsig – OS-ELM using sigmoid activation function
[c] OS-ELMsin – OS-ELM using sine activation function

Table 3
Root-mean squared errors of HTM-MAT and OS-ELM variants using two historical and one real dataset

| Dataset | OS-ELMrbf (RMSE) | OS-ELMsig (RMSE) | OS-ELMsin (RMSE) | HTM-MAT (RMSE) |
|---|---|---|---|---|
| heart_data | 1.5275 | 0.6734 | 1.5816 | 0.2582 |
| australian_data | 0.8236 | 0.7164 | 0.8249 | 0.0381 |
| pressure_data | 0.5075 | 0.5075 | 0.5075 | 0.0000 |

### 4. *Impact*

HTM-MAT is an open source MATLAB toolbox that allows cortical learning algorithms to be explored in a technical computing environment. HTM-MAT to the best of our knowledge presents the first attempt to implement a comprehensive toolbox of HTM cortical like algorithms in a technical computing environment such as MATLAB ®. HTM-MAT is a reusable framework that can easily be adapted to include novel neural-like algorithms that are biologically plausible. This implies that HTM-MAT is indeed a framework for exploring any kind of neural network or technology that is deeply rooted in cortical-like principles. However, this remains an open challenge for neural network experts but a world of possibilities for software developers. It is expected that HTM-MAT will facilitate research in this direction and help advance the state-of-the-art in predictive applications.

### 5. Conclusions

This paper presents HTM-MAT, an open source toolbox implementation of the cortical learning algorithm (CLA) which is based on the hierarchical temporal memory (HTM) theory. HTM-MAT is inspired by how the neocortex operates and proposes an online spatiotemporal sparse-distributed representational neural learning structure. This paper also presents illustrative examples using simple tasks on how to use HTM-MAT. It also compares its prediction performance with results reported using similar applications employing state-of-the-art algorithms on three historical benchmark datasets. The results of simulations show that HTM-MAT is indeed capable of precise word/numeric predictions and can give better results than a popular sequence learning application.